\documentclass[journal,comsoc]{IEEEtran}
\usepackage[T1]{fontenc}
\ifCLASSINFOpdf
\else
\usepackage[dvips]{graphicx}
\fi
\usepackage{url}

\usepackage{graphicx}
\usepackage{epsfig,epstopdf}
\usepackage{algpseudocode}
\usepackage{balance}
\usepackage{amsmath,amssymb,amsfonts}
\usepackage{multicol,multirow}
\usepackage{mathrsfs}
\usepackage{eufrak}
\usepackage{color, graphics,graphicx}
\usepackage{algorithm}
\usepackage{threeparttable}
\usepackage{subfigure,caption}
\usepackage{cancel}
\usepackage{enumitem}
\usepackage{booktabs}
\usepackage{cite}
\usepackage{bm}
\usepackage{amsmath}

\newenvironment*{proof}{\textbf{Proof}\quad}{\hfill $\square$\par}

\definecolor{netone}{RGB}{34,117,220}

\definecolor{nettwo}{RGB}{0,130,75}

\definecolor{qiu}{RGB}{250,10,10}

\begin{document}
	%
	\title{Improving Unsupervised Domain Adaptation by Reducing Bi-level Feature Redundancy}
	
	\title{Improving Unsupervised Domain Adaptation by Reducing Bi-level Feature Redundancy}
	
	\author{Mengzhu~Wang, Xiang~Zhang,~\IEEEmembership{Member,~IEEE,} Long~Lan,~\IEEEmembership{Member,~IEEE,}, \\ Wei Wang,~Huibin~Tan,~Zhigang~Luo~\IEEEmembership{Member,~IEEE,}
		\thanks{This work was supported by the National Natural Science Foundation of China (61806213, 61702134, 61906210)}
		\thanks{M. Wang is with Science and Technology on Parallel and Distributed Laboratory, College of Computer, National University of Defense Technology, Changsha, China. E-mail: dreamkily@gmail.com.}
		\thanks{X. Zhang and L. Lan are with Institute for Quantum and State Key Laboratory of High Performance Computing, National University of Defense Technology, Changsha, China. Email: $\{$zhangxiang08, long.lan$\}$@nudt.edu.cn.}
		\thanks{W. Wang is with the DUT-RU International School of Information Science \& Engineering, Dalian University of Technology, Dalian, Liaoning, 116000, P.R. China e-mail: WWLoveTransfer@mail.dlut.edu.cn}
		\thanks{H. Tan is with the Department of Science and Technology on Parallel and Distributed Processing, National University of Defense Technology, Changsha, China. E-mail: tanhuibin2815@163.com.}
		\thanks{Z. Luo is with Science and Technology on Parallel and Distributed Laboratory, College of Computer, National University of Defense Technology, China. E-mail: zgluo@nudt.edu.cn.(Corresponding to X. Zhang and Z. Luo)}}

	\markboth{IEEE Transactions on Image Processing}
	{Shell \MakeLowercase{\textit{et al.}}: Bare Demo of IEEEtran.cls for IEEE Journals}
	\maketitle

\begin{abstract}
Reducing feature redundancy has shown beneficial effects for improving the accuracy of deep learning models, thus it is also indispensable for the models of unsupervised domain adaptation (UDA). Nevertheless, most recent efforts in the field of UDA ignore this point. Moreover, main schemes realizing this in general independent of UDA purely involve a single domain, thus might not be effective for cross-domain tasks. In this paper, we emphasize the significance of reducing feature redundancy for improving UDA in a \emph{bi-level} way. For the first level, we try to ensure compact domain-specific features with a transferable decorrelated normalization module, which preserves specific domain information whilst easing the side effect of feature redundancy on the sequel domain-invariance. In the second level, domain-invariant feature redundancy caused by domain-shared representation is further mitigated via an alternative brand orthogonality for better generalization. These two novel aspects can be easily plugged into any BN-based backbone neural networks. Specifically, simply applying them to ResNet50 has achieved competitive performance to the state-of-the-arts on five popular benchmarks. Our code will be available at \url{https://github.com/dreamkily/gUDA}.
\end{abstract}


\IEEEpeerreviewmaketitle

\section{Introduction}
\label{sec:intro}

\IEEEPARstart{R}{ecent} attempts on deep neural networks have brought about extraordinary performance in various visual tasks, especially in image classification. Yet, for cross-domain classification tasks, a classifier directly trained on one large annotated dataset (source domain) could degrade on another dataset (target domain) due to the problem of domain drift. A candidate to defeat this issue is domain adaptation~\cite{tzeng2014deep},\cite{sun2016deep}, \cite{long2017deep}, \cite{long2018conditional},\cite{pan2009survey}, which fills the distribution gap between the source domain and target domain. In reality, unsupervised domain adaptation (UDA) could be a promising technique since it does not require the target dataset to be annotated available in the training process. Nonetheless, this also leads to a series of difficulties, of which the most challenging one is how to leverage the unlabeled data from the target domain to reduce domain shift.

The early research efforts in this respect use some proper distance metrics like the maximum mean discrepancy (MMD) \cite{gretton2012kernel} and its variants including maximum mean and covariance discrepancy (MMCD) \cite{zhang2020maximum} to measure inter-domain feature distribution, then trains a model to minimize such distance metrics. Afterwards, a series of attempts \cite{dai2007boosting}, \cite{xu2017unified} are to reweight or select key instances \cite{hubert2016learning} in the source domain to minimize MMD with class-wise information. Other works like deep confusion network \cite{long2017deep} treat MMD and its variants~\cite{saito2018maximum} as a regularization to bound the learned feature distribution across domains. With the advent of generative adversarial networks (GAN) \cite{goodfellow2014generative}, adversarial learning starts to become another main line of confusing feature distribution to learn domain-invariant features. In this sort, a domain-invariant feature generator is trained for two domains to fool a learned discriminator in a two-player game, where the generator learns domain-invariant features, whereas the discriminator helps induce the domain-specific features. For instance, ADDA \cite{tzeng2017adversarial} is the first time to exploit GAN into domain adaptation. CDAN \cite{long2018conditional} considers to simultaneously confuse features and labels across domains by aligning multi-modal structures across domains via multi-linear conditioning strategies. Differently, MADA \cite{pei2018multi} captures multi-modal structures within individual domains by using multiple domain discriminators. Parallel to such studies, another line focuses on exploring network modules like {Batch Normalization} (BN) \cite{ioffe2015batch} for improving UDA. To make BN apply for domain adaptation tasks, AdaBN \cite{li2018adaptive} refines batch normalization to leverage the first- and second-order statistics of the target domain and transfers the knowledge from the source domain to the target domain. Transferable batch normalization \cite{wang2019transferable} delves into the channel-wise transferability of normalization techniques for domain adaptation. Apparently, such preceding novel arts predominantly focus on learning transferable feature representation but few consider whether the learned feature representations contain abundant information and how feature redundancy affects the UDA performance.

As stated in \cite{gu2018recent}, convolutional neural networks (CNNs) have significant redundancy between filters and feature channels. This, in a sense, obstructs network performance gains  for the imbalanced distribution of spectrum. Recently, promoting diversity via weight orthogonality~\cite{bansal2018can}, and decorrelating features~\cite{roy2019unsupervised} can make feature spectrum near uniform in different manners. Nevertheless, little progress explicitly applies such insights for UDA. Importantly, such insights are in general designed for single-domain visual tasks and could not be suitable for cross-domain counterparts. To this end, we revisit the UDA problem from the perspective of feature redundancy and propose to prevent feature redundancy for improving UDA models in a \emph{bi-level} way. Specifically, we first explore a transferable decorrelated batch normalization module (TDBN) to prevent domain-specific feature redundancy by jointly learning transferable and decorrelated features. We then explore a novel orthogonality based regularization for enhancing the diversity of domain-invariant features to mitigate the corresponding feature redundancy. As a result, by simply plugging two components into a pre-trained BN-based backbone ResNet50, a deep UDA model is constructed to simultaneously reduce the redundancy of both intra-domain and inter-domain features, which could ease overfitting and achieve better generalization. Through comprehensive experiments on five widely-used UDA benchmarks, reducing the bi-level feature redundancy shows the promising efficacy for UDA, and the corresponding deep UDA model yields encouraging performance as compared to several state-of-art siblings.

Our main contributions are summarized as follows:

$\bullet$ We provide a novel perspective on how to reduce feature redundancy for improving UDA. A \emph{bi-level} way to reduce feature redundancy is proposed, which can easily be well coupled with any BN-based backbone networks for UDA.

$\bullet$ A transferable decorrelated normalization module (TDBN) is devised to prevent domain-specific feature re\textbf{}dundancy through relieving intra-domain feature co-adaptation for better generalization.

$\bullet$ A novel orthogonal regularization is proposed to enhance inter-domain feature diversity, which can serve as an alternative approach to realize the orthogonality. It helps to reduce domain-invariant feature redundancy by stabilizing the distributions of features as well as regularizing the networks to induce compact features. Besides, it yields (near) orthogonality without the usage of singular value decomposition (SVD).

 {\section{Related Work}} \label{2}
 In this section, we will review some recent trends in unsupervised domain adaptation that is mostly related to our approach.\\
 
\subsection{Unsupervised Domain Adaptation}
The goal of UDA is to transfer knowledge \cite{yu2017transfer}, \cite{jing2020adaptive}, 
\cite{Chang_2019_CVPR} from an annotated source domain to another unlabeled target domain by reducing domain shift. From the standpoint of feature learning, many UDA studies can be considered as either domain-invariant feature learning or domain-specific feature learning.

\textbf{Domain-invariant feature learning}. In UDA, the primary pursuit is to learn domain-invariant features. Many mainstream approaches \cite{Yan_2017_CVPR}, \cite{zellinger2017central}, \cite{zhang2020maximum} belong to this category of domain-invariant feature learning and achieve this pursuit by aligning data distributions across domains in an either coarse or fine-grained manner.

The idea behind coarse alignment approaches are the usage of either global distribution alignment or global data structure. For global distribution alignment, many distribution discrepancy metrics like WMMD \cite{Yan_2017_CVPR}, CMD \cite{zellinger2017central}, and MMCD \cite{zhang2020maximum}, which consider data statistics information, are directly minimized by those UDA models. Correlation alignment (CORAL) \cite{sun2016deep} minimizes the discrepancy between the covariance matrices of the whole source features and the target features. Wasserstein Distance (WDGRL) \cite{shen2017wasserstein} learns domain-invariant representation to reduce domain shift based on Wasserstein distance. Also, adversarial learning has been widely used for coarse domain alignment. Domain adversarial neural networks (DANN) \cite{ganin2016domain} confuses domain features via domain adversarial loss to induce domain-invariant representation. Joint adaptation network (JAN) \cite{long2017deep} incorporates MMD \cite{gretton2012kernel} with adversarial learning to align the joint distributions based on specific layers across domains. Besides, deep relevance network considers global low-rank structure to align different domains via discriminative relevance regularization (DRR) \cite{zhang2020enhancing}.
For fine-grained alignment, such efforts seek to match multimode data structures like class-wise margin and data distributions \cite{pei2018multi}, \cite{long2018conditional}, \cite{hoffman2018cycada}, \cite{jingjing2020maximum}. Multi-adversarial domain adaptation (MADA) \cite{pei2018multi} captures the multimode structures to enable fine-grained alignment of different data distributions based on local domain discriminators to learn domain-invariant feature. Conditional domain adaptation network (CDAN) \cite{long2018conditional} simultaneously aligns features and labels and overcome mode mismatch in adversarial learning for domain invariance. Cycle-consistent adversarial domain adaptation (CyCADA) \cite{hoffman2018cycada} introduces a cycle-consistency loss to ensure the model the consistency of local semantic mapping.

\textbf{Domain-specific feature learning}. Domain-specific feature learning assumes that varied domains have different information, which should be separated from domain-invariant features. In general, methods in this respect consider to devise network components and loss function for each domain. Of them, Batch Normalization (BN) is a versatile network component that has been extended to learn domain-specific features for UDA \cite{Chang_2019_CVPR}, \cite{wang2019transferable}. DSBN \cite{Chang_2019_CVPR} adapt to both domains by specifying batch normalization layers to separate domain-specific information from domain-invariant features in an explicit manner. Transferable Normalization (TransNorm) \cite{wang2019transferable} associates domain-specific features via the attention mechanism to realize transferability. Adaptive Batch Normalization (AdaBN) \cite{li2018adaptive} proposes a brute-force post-processing method to replace batch normalization statistics with the counterparts of target samples. Domain-specific whitening transform (DWT) \cite{roy2019unsupervised} proposes domain-specific alignment layers to align domain-specific covariance matrices of intermediate features. Different from them, which only encourage the transferability across domains, this work focuses on the reduction of feature redundancy for better generalization. Besides, our method learns domain-specific features and domain-invariant features, simultaneously, and enjoys both benefits.

\subsection{Generalization Strategies for DNNs}
Many approaches seek to improve DNNs for better generalization by easing overfitting. Towards this goal, several regularizers and network structures have been developed to improve model generalization ability. For instance, early stopping \cite{yao2007early} achieves this aim by keeping from abusing finite samples as early as possible. Weight decay \cite{krogh1992simple} penalizes large weights in stochastic gradient descent (SGD) like $\mathcal{L}_{2}$ regularization to reduce model complexity. Decoupled weight decay regularization \cite{loshchilov2017decoupled} further extends weight decay to be applicable for Adam by decoupling weight decay from the gradient-based update. Model ensemble \cite{wang2003mining} is a kind of effective methods for model generalization by averaging the inference results of multiple models. Of them, Adaboost \cite{vezhnevets2005modest} is a basic and versatile form, which concatenates a group of weak classifiers to yield a united stronger classifier. Data augmentation \cite{reed1992regularization} augments different variations about a few instances by hand to help the model to learn invariant features and behaves like a regularizer, thereby easing overfitting. Dropout \cite{srivastava2014dropout}, \cite{srinivas2016generalized} can also regularize the network for sake of the behavior of the model ensemble. Different from Dropout, DropConnect \cite{wan2013regularization} tries to randomly drop a portion of the weights rather than feature activations to regularize the network in an implicit way. BN \cite{ioffe2015batch}, \cite{kohler2018towards} can also handle the internal covariate shift problem to stabilize the network training. To reduce feature redundancy, decorrelated batch normalization \cite{huang2018decorrelated} takes a step to promote generalization by decorrelated learning. Besides, orthogonality \cite{bansal2018can} stabilizes the distribution of the network activation with efficient convergence and achieves better generalization. Such techniques reported in the literature have shown the ability to improve network generalization ability. Motivated by such insights, we utilize  decorrelated learning and orthogonality to improve the model of UDA from the view of redundancy minimization.

\subsection{Orthogonality in DNNs}
As claimed in \cite{bansal2018can}, \cite{huang2017orthogonal}, \cite{yang2019mean}, orthogonality imposed over the weights is able to stabilize the optimization of DNNs by preventing the explosion or vanishing of back-propagated gradients. In \cite{wan2013regularization}, the orthogonality is applied in singal processing since it is capable of preserving activation energy and reducing redundancy in representation. Yang \textit{et al}. \cite{yang2019mean}, \cite{xie2017all} proposed that orthogonal weight normalization in the nonlinear sigmoid network can obtain dynamical isometry. Huang \textit{et al}. \cite{huang2017orthogonal} showed a novel orthogonal weight normalization method can guarantee stability and formulate this problem in feed-forward neural networks. Also, Huang \textit{et al}. \cite{huang2020controllable} proposed a computationally efficient and numerically stable orthogonality method to learn a layer-wise orthogonal weight matrix in DNNs, which enables to control the orthogonality of a weight matrix. There is also some research in Recurrent Neural Networks (RNNs) based on orthogonal matrices to avoid the gradient vanishing problems. Different from above, our method is related to the methods that impose orthogonal regularization in loss function under Frobenius norm, it can stabilize training via ensuring fully-connected layers output to be identical distributions which reduce the feature co-adaptation. Orthogonal to such siblings, the proposed (near) orthogonality is implicitly obtained through the joint matrix trace and determination constraints in theory. To the best of our knowledge, this orthogonality is first mentioned here and applied for the UDA problem.

\section{Method}
This section reconsiders the UDA problem from the view of feature redundancy and readily details how to reduce bi-level feature redundancy for improving unsupervised domain adaptation model (UDA).

\subsection{Preliminary}
In this paper, we will focus on the UDA, where the source domain has enough labels whereas target domain has unlabeled examples. In UDA, we suppose $D_{s}=\left\{\left(x_{i}^{s}, y_{i}^{s}\right)\right\}_{i=1}^{n_{s}}$ and $D_{t}=\left\{x_{j}^{t}\right\}_{j=1}^{n_{j}}$ to be the labeled source data and unlabeled target, drawn from different distributions respectively. The data probability distributions of $X_{s}$ and $X_{t}$ are $P$ and $Q$, respectively, where $X_{s} \sim P$ and $X_{t} \sim Q$, in UDA we have $P \neq Q$. Our goal is to predict the target label and reduce the difference between the two distributions.
\subsection{Model}
\begin{figure*}[t]
	\centering
	\includegraphics[width=14cm, height=6cm]{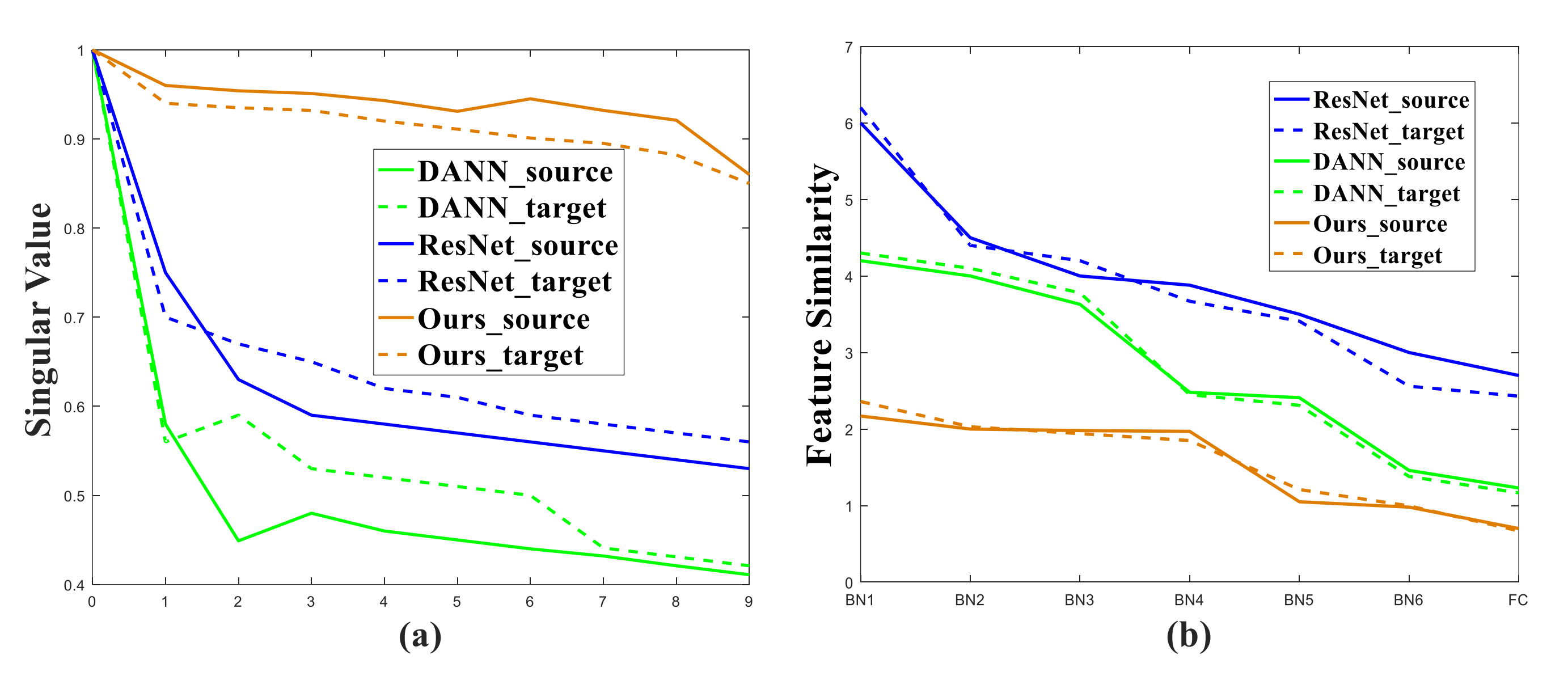}
	\caption{The distribution of feature spectrum and layer-wise feature similarity of ResNet50, DANN and our UDA model, respectively: the proposed bi-level way to reduce feature redundancy definitely makes the feature spectrum near uniform as compared to both ResNet50 and DANN. Note that, for the results of our model, the notations BN1, $\cdots$, BN6 for concise are abused here but they actually correspond to the proposed BN modules. In ResNet50 and DANN, they stand for vanilla batch normalization module.} 
	\label{discrepancya}
\end{figure*}

\textbf{Motivation}. In the field of UDA, the mainstream methods focus on filling the distribution gap between the source and the target, due to domain shift. As a matter of fact, we easily neglect the inherent vulnerability of deep networks in themselves such as the instability in network learning and the redundancy of feature representation. Apparently, such internal weaknesses could continue to deteriorate in cross-domain tasks for the sake of cross-domain data heterogeneity. Nevertheless, few efforts truly discuss the effect of this insight in domain adaptation; in contrast, most arts consider to learn discriminative or transferable feature patterns for cross-domain classification tasks. Different from such arts, this study primarily aims to prevent feature patterns from being redundant, namely, reducing feature redundancy or redundancy minimization. 

Recent studies \cite{wang2020orthogonal} claim that the imbalanced distribution of the feature spectrum could contribute to feature redundancy, and fortunately, feature diversity or feature decorrelation can mitigate this realistic issue to some degree. Besides, many works judge feature redundancy with feature similarity. Inspired by such previous insights, we start to scrutinize whether there exists feature redundancy in existing UDA models, in light of both the distribution of the feature spectrum and layer-wise feature similarity. As Fig.~\ref{discrepancya} shows, the distribution of feature spectrum in both ResNet50 and domain adaptation neural networks (DANN) on the transfer task {$\mathbf{A}\rightarrow\mathbf{C}$} of the Office-Home dataset are some imbalanced, and meanwhile their corresponding feature similarities, coming from the channels in different BN layers and features in Fc layers, are still highly similar. Especially, DANN has better transferability with lower feature similarity than ResNet50, but still suffers from relatively serious distribution imbalance of spectrum. This implies that low feature similarity is not enough to prevent feature redundancy. Instead, ResNet50 has weaker transferability with higher feature similarity but has relatively balanced distribution of spectrum as compared to DANN. This also indicates that the near uniform distribution of the sepctrum is still insufficient to ensure redundancy minimization as well. Based on these discussions, we consider to explicitly reduce feature redundancy in a \emph{bi-level} way, which jointly make feature similarity much lower and distribution of feature spectrum close to uniform. As in Fig.~\ref{discrepancya}, in contrast with both ResNet50 and DANN, the proposed approach plugged into ResNet50 can significantly reduce layer-wise feature similarity of ResNet50 as well as mitigate the corresponding imbalanced feature spectrum. Generally speaking, making feature spectrum near uniform is beneficial for stabilizing network learning and model convergence, while reducing feature similarity assists in achieving better model generalization ability. Since the proposed method can achieve both goals--\emph{one stone two birds}, our model is definitely helpful for reducing feature redundancy to promote UDA performance.

Build off the above analysis, the proposed approach to reducing feature redundancy contains two aspects or levels: the first level is designed for the reduction of domain-specific (or intra-domain) feature redundancy, while the second seeks to prevent domain-invariant (or inter-domain) feature redundancy. Fig.~\ref{discrepancya}(b) shows that, in both ResNet50 and DANN, the domain-specific features learned by BN layers and domain-invariant features by Fc layers are to a certain redundant, while our results have lower feature similarity and near uniform distribution of feature spectrum. Thus, this \emph{bi-level} manner is  feasible. In detail, we at first try to learn as compact domain-specific features as possible through a transferable decorrelated normalization module (TDBN), with the goal of maintaining specific domain information as well as reducing the risk of this feature redundancy on the following domain invariance. Then, domain-invariant feature redundancy caused by domain-shared representation can be further reduced by using a novel orthogonality for feature diversity. The above bi-level scheme is lightweight and can be easily plugged into ResNet50 as our UDA model (Ours in short). As empirical studies show, the proposed UDA model \emph{without} extra hassle can achieve competitive performance with several state-of-the-arts.

\begin{figure*}[t]
	\centering
	\includegraphics[width=17cm]{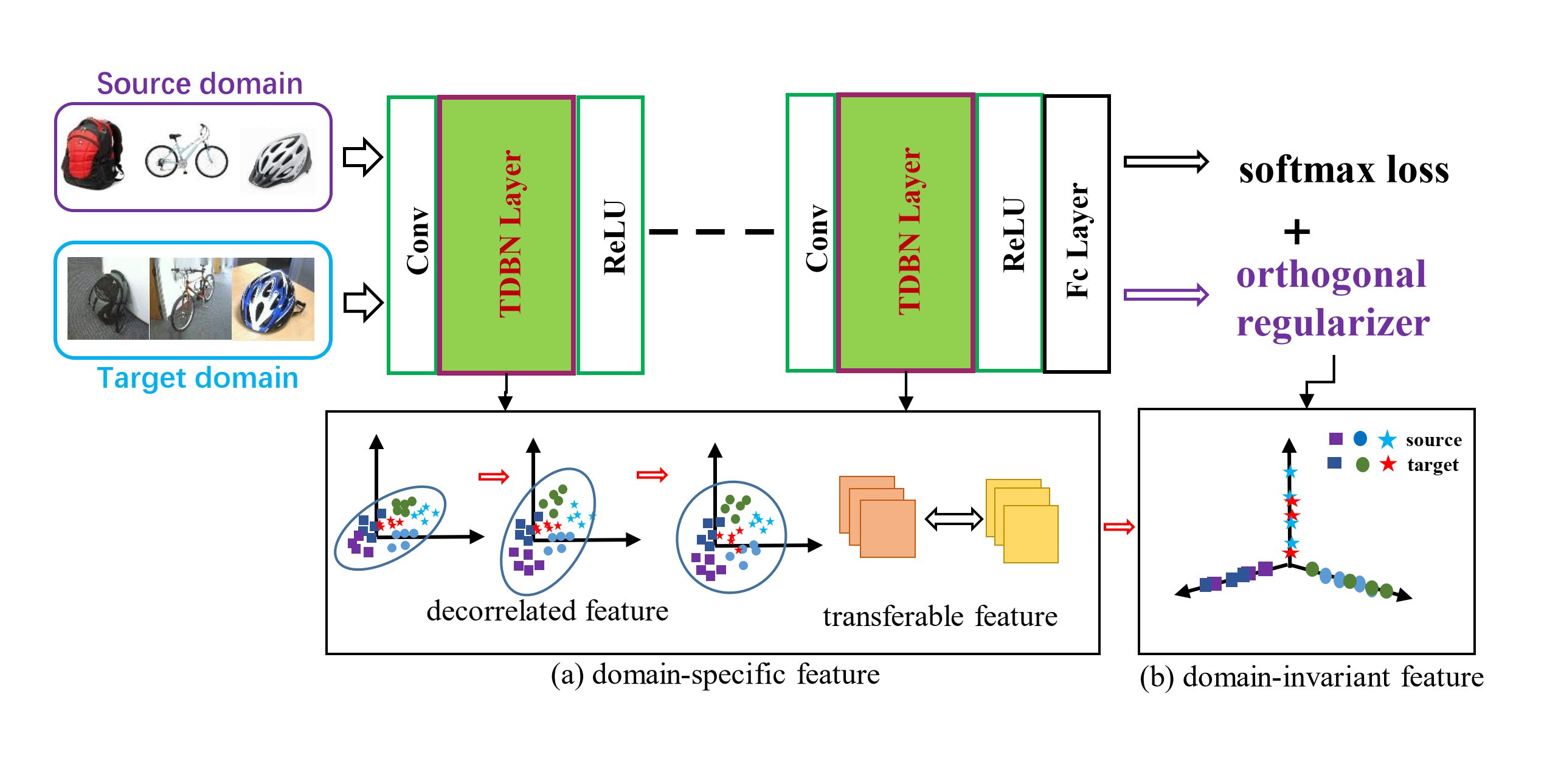}
	\caption{Illustration of the proposed UDA model based on ResNet50 with a \emph{bi-level} way to prevent feature redundancy: (a) each BN is replaced with TDBN, for the purpose of reducing domain-specific feature redundancy; (b) the pre-softmax weights in fully-connected layer are constrained to be (near) orthogonal by an orthogonal regularizer, which keeps domain-invariant features from being redundant.} \label{illustration}
\end{figure*}

\textbf{The Flowchart}. For clarity, the proposed UDA model is illustrated in Fig.~\ref{illustration}. To hinder the overfitting risk of domain-specific feature redundancy, TDBN first mitigates feature co-adaptation and then starts to proceed feature transferability. Different from existing BN-based modules, TDBN not only reduces feature redundancy but also learns transferable features, without introducing any extra parameters. Albeit simple, the performance gain is attractive. For the ultimate classifier in the Fc layer, the pre-softmax weights are imposed to be (near) orthogonal for domain-invariant feature diversity, thereby reducing feature redundancy. This is because the orthogonality can extract the more compact feature bases to span the whole feature subspace. Thus, it encourages to learn compact features. This property is so amazing to reduce the risk of model overfitting as well. Besides, the orthogonality still shares extra appealing properties: accelerating model convergence and stabilizing network training. The difference from existing orthogonalities lies in that the proposed orthogonality does not involve either singular value decomposition (SVD) or the iterative calculation of the largest and smallest singular values.

\subsection{Reducing Domain-specific Feature Redundancy}
Recall that domain-specific feature learning assumes that each domain has distinct information, thus domain-specific features should be separated from domain-invariant features. In the field of UDA, batch normalization (BN) has been extended to learn domain-specific features, which calculates domain-independent statistical information, i.e., the mean and variance, for standardization. It serves as an important network component to accelerate network convergence by reducing the co-variant shift. Most BN-based modules are based on standardization, which have no ability to simultaneously reduce feature redundancy and realize feature transferability. Towards this goal, we couple the decorrelated BN \cite{huang2018decorrelated} with the channel-wise transferability attention idea from transferable BN \cite{wang2019transferable}, to jointly enjoy their strengths: the former mitigates feature co-adaptation to reduce feature redundancy, while the latter focuses on feature transferability across domains. For completeness, we detail our BN module termed Transferable Decorrelated Batch Normalization (TDBN) as below.

In a BN module, given a mini-batch input, the corresponding standard normalized outputs become:

\begin{align} \label{eq:weight}
{\hat{x}_{i}=\gamma \frac{x_{i}-\mu}{\sqrt{\sigma^{2}+\varepsilon}}+\beta},
\end{align}
where $\mu=\frac{1}{m} \sum_{j=1}^{m} x_{j}$ and $\sigma^{2}=\frac{1}{m} \sum_{j=1}^{m}\left(x_{j}-\mu\right)^{2}$ are the mean and variance of the mini-batch,  
$\varepsilon>0$ is a predefined small number to prevent numerical instability, both $\gamma$ and $\beta$ are extra learnable parameters, $m$ is the number of the mini-batch samples.

BN merely performs feature standardization, without conducting feature decorrelation, thus it does not reduce feature redundancy. Huang \textit{et al}. \cite{huang2018decorrelated} proposed a Decorrelated Batch Normalization (DBN) module to attack this issue, and then proceeded to improve DBN with the iterative optimization algorithm for efficiency \cite{Huang_2019_CVPR}. Intrinsically, they contribute to doing ZCA whitening on the standardized inputs. Thus the concrete formula of this process is:
\begin{align}
\textit{BNW}(X; \Sigma) 
&= \gamma \hat{X}   + \beta, \label{eq.scale-shift-after-whitening} \\
\hat{X}_i &= D \Lambda^{-\frac{1}{2}} D^{T}\left(X-\mu \cdot 1^{T}\right), \label{eq.WC-whitening}
\end{align}
where the covariance matrix is
 \begin{align}
 \Sigma=\frac{1}{m}\left(X-\mu \cdot 1^{T}\right)\left(X-\mu \cdot 1^{T}\right)^{T}+\varepsilon I, \label{eq:covariance matrix}
 \end{align}
 and the corresponding singular value decomposition (SVD) form $\Sigma=\mathrm{D} \Lambda \mathrm{D}^{T}$, wherein $\Lambda=\operatorname{diag}\left(\sigma_{1}, \ldots, \sigma_{d}\right)$ and 
$\mathrm{D}=\left[\mathrm{d}_{1}, \ldots, \mathrm{d}_{d}\right]$ are the singular values and the singular vectors, respectively. Iterative whitening normalization efficiently computes $\Sigma^{-\frac{1}{2}}$ by the newton iteration method. $\Sigma^{-\frac{1}{2}}$ can be calculated as follows:
\begin{equation}
\label{eq:sigman}
\begin{array}{c}
\Sigma_{N}=\Sigma / tr(\Sigma),
\end{array} 
\end{equation}
where $tr(\Sigma)$ indicates the trace of $\Sigma$, then $\Sigma_{N}^{-\frac{1}{2}}$ can be calculated as follows:
\begin{equation}
\label{eq:sigma}
\begin{array}{c}
\Sigma^{-\frac{1}{2}}=\Sigma_{N}^{-\frac{1}{2}} / \sqrt{t r\left(\Sigma\right)}.
\end{array} 
\end{equation}

Whitening the activation ensures that all the dimensions along the eigenvectors have equal importance in the subsequent linear layer, while standardization ensures that the normalized output gives equal importance to each dimension by multiplying the diagonal scaling matrix. Obviously, it ignores the interplay between the source domain and the target domain. Similar to TransNorm, which tries to align the sufficient statistics of both domains via channel transferability, we also quantify the transferability of different channels and adapt them for different domains. For each channel $j$ and the whiten data $x$, the importance of the different channels can be determined by:
\begin{equation}
\label{eq:alpha}
\begin{array}{c}
\alpha^{(j)}=\frac{c\left(1+d^{(j)}\right)}{\sum_{k=1}^{c}\left(1+d^{(k)}\right)}, j=1,2, \ldots, c,
\end{array} 
\end{equation}
where 
\begin{equation}
\label{eq:d}
\begin{array}{c}
d^{(j)}=\left|\frac{\mu_{s}^{(j)}}{\sqrt{\sigma_{s}^{2(j)}+\varepsilon}} \cdot \frac{\mu_{t}^{(j)}}{\sqrt{\sigma_{t}^{2(j)}+\varepsilon}}\right|, j=1,2, \ldots, c,
\end{array} 
\end{equation}
wherein $c$ denotes the number of channels in the layer. In contrast to TransNorm \cite{wang2019transferable}, the distance has been replaced by the similarity without introducing extra parameters. Following \cite{wang2019transferable}, we get the final output of the proposed module TDBN as below:

\begin{align}
y_{s}&=(1+\alpha)\left(\gamma x_{s}+\beta\right), \label{eq:ys} \\
y_{t}&=(1+\alpha)\left(\gamma x_{t}+\beta\right). \label{eq:yt}
\end{align}

\subsection{Preventing Domain-invariant Feature Redundancy}\label{Sec_domain}
Learning domain invariant features means features with transferability across domains, which is always a predominantly expected aim of UDA. Hence, it plays a critical role in the ultimate inference ability of the classifier on the target domain. Considering this, we try to constrain the pre-softmax weights in the Fully-connected (Fc) layer, which behaves like a classifier, to induce domain invariance. Recall that increasing feature diversity can assist in preventing feature redundancy, while the orthogonality just has the ability to cope with this issue. Thus, we explore an orthogonal regularization over the pre-softmax weights of the classifier for feature diversity.

As one knows, the orthogonality has shown great success in improving the stability of deep neural networks, because it can preserve energy, stabilize activation distributions, and ensure uniform spectrum. Due to such appealing properties, many studies explore the orthogonality \cite{bansal2018can}, \cite{huang2017orthogonal}, \cite{huang2020controllable}, \cite{jia2019orthogonal} in different ways to promote DNNs. Some studies regard it as the optimization problem on Stiefel Manifolds \cite{balogh2005global}, while the others consider its soft version as a differentiable regularizer \cite{bansal2018can} to regularize CNNs. Such novel insights spur us to develop another alternative orthogonal regularizer by avoiding iterative spectrum computation for efficiency. Moreover, it can guarantee the singular values to be near ones in theory. This significantly differs from most preceding arts \cite{bansal2018can}.

Prior to detailing our orthogonality, we at first present its constraint form of our orthogonality as below:
\begin{equation}
\label{eq:loss}
\begin{array}{c}
loss = {\min _\theta }{\mathbb{E}_{(x,y) \in D}}[\mathcal{L}(y,f(x;\theta ))]\\
s.t.\left\{ {\begin{array}{*{20}{c}}
	{\det (W{W^T}) = 1}\\
	{\operatorname{trace}(W{W^T}) = n}
	\end{array}} \right.{\rm{ }},
\end{array} 
\end{equation}
where $W$ is the pre-softmax weight matrix. Note that our orthogonal constraint is not limited to the pre-softmax weights. $det(\bullet)$ denotes the determination of the square matrix, while $trace(\bullet)$ is the trace of the square matrix. This constraint can implicitly ensure the weight matrix to be orthogonal in theory.

\textbf{Lemma 1.} Any non-singular symmetric matrix $G \in R^{n \times n}$ has the singular values of all the ones under the orthogonal constraint of Eq.~\eqref{eq:loss}.

\begin{proof}
By simple algebra, one knows:
\begin{equation}
\label{eq:trace}
\begin{array}{c}
\begin{array}{l}\operatorname{trace}(G)=\sum_{i} \lambda_{i}, \\ \operatorname{det}(G)=\prod_{i} \lambda_{i}\end{array} i=1, \ldots, n,
\end{array} 
\end{equation}
where $\lambda_{i}$ is the $i$-th singular value of $G$. According to Eq.~\eqref{eq:trace}, the inequalities hold:
\begin{equation}
\label{eq:eqa11}
\begin{array}{c}
\sum_{i} \lambda_{i} \geq n, \\
\prod_{i} \lambda_{i}=1.
\end{array} 
\end{equation}
 
 Of them, the first inequalities of \eqref{eq:eqa11} contain the constraint Eq.~\eqref{eq:loss}, thus it naturally holds. According to the basic inequality theory given in \textbf{APPENDIX}, when $\lambda_{i} \geq 0, i=1, \cdots, n$ there holds:
 \begin{equation}
 \label{eq:eqa12}
 \begin{array}{c}
\lambda_{1}=\lambda_{2}=\cdots=\lambda_{n}.
 \end{array} 
 \end{equation}

This completes the proof.
\end{proof}

\textbf{Theorem 1.} The learned weight matrix $W \in R^{n \times m}$ under the constraint of Eq.~\eqref{eq:loss} could be orthogonal for $n<<m$.

{\bf Proof.} Assume that $G=W W^{T}$, where $n<<m$. Then, $G$ is the most likely to be a non-singular symmetric matrix. In practice, this point can be guaranteed. Assume that there exists the singular value decomposition of $W$, i.e., $W=U \Sigma V^{T}$. Accordingly, $G=U \Sigma U^{T}$. According to \textbf{Lemma 1}, $G=U U^{T}$ because $\Sigma=I$. Since $U$ is a square matrix corresponding to all the singular values, there still holds $G=I$. This implies that $W W^{T}=I$.

To our best knowledge, no previous works explicitly mention these findings. Besides, it is non-trivial to directly optimize the above constrained objective. Thus, we treat this orthogonal constraint as a regularization for efficient optimization. This type of relaxation is also the mainstream of implementing orthogonality. Following this, we rewrite Eq.~\eqref{eq:eqa12} as: 
 \begin{equation}
\label{eq:eqa13}
\begin{array}{c}
loss =\min _{\theta} \mathbb{E}_{(x, y) \in D}[\mathcal{L}(y, f(x ; \theta))] \\
+ \beta\left(\left\|\operatorname{det}\left(W W^{T}\right)-1 \right\|^{2}+\left\|\operatorname{trace}\left(W W^{T}\right)-n\right\|^{2}\right)
\end{array} 
\end{equation}
where $\beta$ is a regularization parameter. We again resort to auto-computing the gradient of Eq.~\eqref{eq:eqa13} for simplicity. We do not need to compute the expensive SVD or iteratively calculate singular values such as the regularizer based on spectral restricted isometry property \cite{baraniuk2008simple}. Our orthogonality in an implicit fashion encourages the network to learn diverse features so that features become more compact.

\section{Experiments}\label{experiment}
\begin{figure*}[t]
	\centering
	\includegraphics[width=13cm, width=11cm]{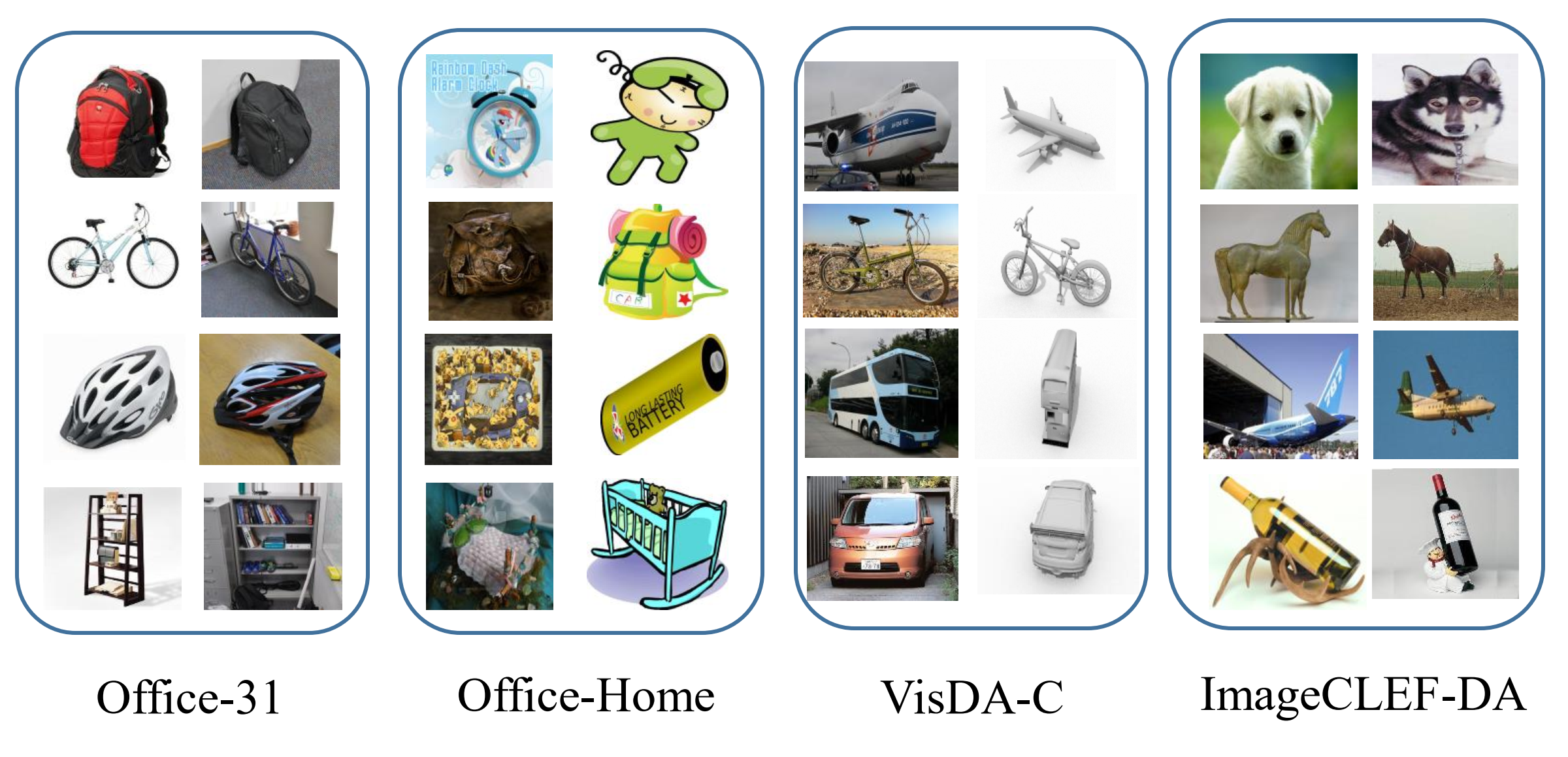}
	\caption{Object recognition sample images from each dataset. (a) Office-31 dataset images on three domains, (b) Office-Home dataset images on four domains, (c) VisDA-C dataset images on twelve domains, and (d) ImageCLEF-DA dataset images on twelve domains} 
	\label{framework}
\end{figure*}
\begin{figure}[t]
	\vspace{0.1in}
	\centerline{\includegraphics[width=8cm]{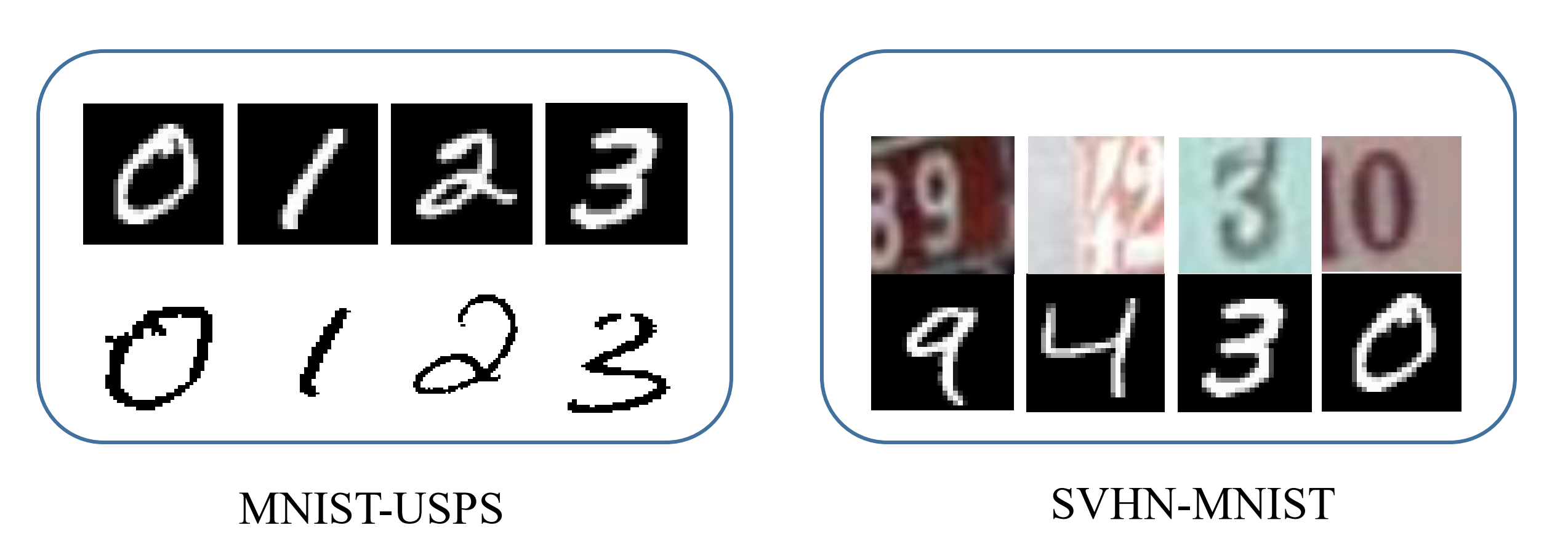}}
	\vspace{0.1in}
	\caption{(a) MNIST and USPS dataset images on ten domains, (b) MNIST and SVHN dataset images.}
	\label{difference}
\end{figure}

This section will detail the implementation of our UDA model (Ours in short) and the corresponding training protocols, and report our experimental evaluation in the end. We conduct five experiments on both small and large-scale datasets to evaluate our method as compared with its well-behaved siblings. Moreover, we also present an ablation study to analyze the effect of each component over UDA.
\subsection{Experimental Settings}
\textbf{Data Preparation.}
We conduct five experiments on seven datasets: Office-31, Office-Home, ImageCLEF-DA, MNIST, USPS, SVHN, and VisDA-C. These datasets as popular benchmarks have been widely adopted to evaluate domain adaptation algorithms in previous arts. 

{\bf Office-31}~\cite{saenko2010adapting} dataset is a standard benchmark for testing domain adaptation methods. It contains 4,652 images organized in 31 classes from three different domains: Amazon (A), Dslr (D), Webcam (W). Amazon images are collected from amaon.com, while both Webcam and Dslr images are manually collected in an office environment, we evaluate all the compared methods on all six transfer tasks including  A$\rightarrow$W, A$\rightarrow$D,
 D$\rightarrow$W, W$\rightarrow$A, D$\rightarrow$A, and W$\rightarrow$D.

{\bf Office-Home}~\cite{venkateswara2017deep} dataset contains four distinct domains including Art (A), Clipart (Cl), Product (Pr), and Real World(Rw), and each domain involves 65 different categories. There are 15,500 images totally in the dataset, thus it serves as a large-scale benchmark for testing domain adaptation methods.

{\bf ImageCLEF-DA}\footnote{http://imageclef.org/2014/adaptation} dataset is a benchmark for ImageCLEF 2014 domain adaptation, which is composed of 12 common categories shared by three public datasets. They are Caltech-256 (C), ImageNet ILSVRC 2012 (I), and Pascal VOC 2012 (P), respectively. There have 50 images category and 600 images in each domain. We evaluate the methods on all the six transfer tasks: C$\rightarrow$I, C$\rightarrow$P, P$\rightarrow$I, P$\rightarrow$C, I$\rightarrow$C, and I$\rightarrow$P.

{\bf VisDA-2017}~\cite{peng2017visda} is a challenging simulation-to-real dataset, it has two very distinct domains: synthetic, renderings of 3-D models from different angles and with different lighting conditions, and real natural images. It contains 12 classes in the training, validation, and test domains

{\bf MNIST (M)} dataset comes from the National Institute of Standards and Technology. It is a standard digit recognition datasets covering ten handwritten digits from 0-9 in the form of $28 \times 28$ pixels grayscale images. It is divided into both training and testing sets. The training dataset comes from 250 different people, of which 50\% are from high school students, and the remaining from Census Bureau workers; the testing dataset is similar to training dataset. This dataset contains 60,000 training images and 10,000 testing images. Similar to {\bf MNIST (M)}, {\bf USPS (U)} also contains 10 classes from 0-9, and has 7,291 training and 2,007 testing images. All the images are cropped as $16 \times 16$ grayscale pixels. It is subject to significantly different probability distributions to MNIST. Besides, {\bf SVHN (S)} \cite{netzer2011reading} contains colored 10-digit images with the size 32$\times$32, which contains 73,257 training images and 26,032 testing images.

\textbf{Implementation Details}.
For fairness, in object recognition tasks, we adopt the same base network ResNet50 as our backbone. For example, on Office-Home dataset, we use ResNet50 \cite{he2016deep} as our backbone. Especially for VisDA-2017, owing to its huge volume, we use ResNet101 as our backbone. We will fine-tune all convolutional and pooling layers from the ImageNet pretrained models and then train the classifier layer by back-propagation. This network is initialized with weights from ImageNet dataset. For our model, we also adopt the same configurations. In the fully connected layers, we discard original class numbers and change class numbers to 65, and this matches the Office-Home classes, Also, we will discard original class numbers and change class numbers to corresponding domains. During the training phase, we set the mini-batch size to 24, and the epoch to 200. We use Adam optimizer with an initial learning rate of 0.01 and the weight decay is $5 \times 10^{-4}$. The learning rate will decrease by a factor of 10 after 50 and 90 epochs. The maximum number of the iterations of the optimizer is set to 10,000. For digit classification, we follow the protocols in \cite{tzeng2017adversarial}. We set the hyper parameter $\beta=1$ for all the tasks.

We implement the proposed UDA model in PyTorch, and report the average classification accuracy on both object recognition and digit classification. On the Office-31 dataset, we compare our method with popular transfer learning and recent transfer learning methods including deep residual learning for image recognition (ResNet50) \cite{he2016deep}, Geodesic flow kernel for unsupervised domain adaptation (GFK) \cite{gong2012geodesic}, Domain-adversarial training of neural networks (DAN) \cite{ganin2016domain}, Joint adaptation networks (JAN) \cite{long2017deep}, Multi-adversarial domain adaptation (MADA) \cite{pei2018multi}, Collaborative and adversarial network for unsupervised domain adaptation (iCAN) \cite{zhang2018collaborative}, Conditional adversarial domain adaptation (CDAN and CDAN+E) \cite{long2018conditional},Confidence regularized self-learning\cite{zou2019confidence},Batch nuclear-norm maximization under label insufficient situations~\cite{cui2020towards}, and the results of the other compared methods are from the original paper for a fair comparison. On the Office-Home, we compare our UDA model with ResNet-50, DAN, JAN, Unsupervised domain adaptation using feature-whitening and consensus loss (DWT and DWT-MEC) \cite{roy2019unsupervised}, and compare it with ResNet101, DAN, deep adversarial neural networks (DANN), maximum classifier discrepancy (MCD) \cite{saito2018maximum}, CDAN, Batch spectral penalization (BSP+DANN and BSP + CDAN) \cite{chen2019transferability} and Domain-Specific Batch Normalization for Unsupervised Domain Adaptation(DSBN) \cite{chang2019domain} on VisDA-2017, respectively. On ImageCLEF-DA, we compare our UDA model with several standard deep learning methods and deep transfer learning methods: ResNet50, DAN, DANN \cite{ghifary2016deep}, JAN, CDAN, and CDAN + E, CAN and iCAN, Visual Domain Adaptation with Manifold Embedded Distribution Alignment (MEDA), Adversarial Tight Match (ATM)~\cite{jingjing2020maximum},DSAN,Self-adaptive re-weighted adversarial domain adaptation~\cite{wang2020self}. For digit classification, we compare our model with CORAL, Simultaneous deep transfer across domains and tasks (MMD) \cite{tzeng2015simultaneous}, Visual Domain Adaptation with Manifold Embedded Distribution Alignment (DRCN) \cite{wang2018visual}, Coupled generative adversarial networks (CoGAN) \cite{liu2016coupled}, Adversarial discriminative domain adaptation (ADDA) \cite{tzeng2017adversarial}, Unsupervised image-to-image translation networks (UNIT) \cite{liu2017unsupervised}, Asymmetric Tri-training for unsupervised domain adaptation (ATT) \cite{saito2017asymmetric}, Generate to adapt: Alignment domains using generative adversarial net (GTA) \cite{sankaranarayanan2018generate}, and Learning Semantic Representations for Unsupervised Domain Adaptation (MSTN) \cite{xie2018learning}, while the results of baselines are borrowed from \cite{roy2019unsupervised}.

\begin{table}[t]
	\centering

	\footnotesize
	\caption{Accuracy (\%) of the digit recognitions. The best results are highlighted by bold numbers.}
	\label{tab:digit}
	\footnotesize
	\vspace{5pt}
	\begin{tabular}{lccc}
		\toprule
		Method  &M$\rightarrow$U& U$\rightarrow$M  & S$\rightarrow$M \\
		\midrule
		Source Only & $89.9$ & $ 57.1$ & $60.1$\\ 
		MMD (2015 ICCV)& $81.1$ & $-$ & $71.1$\\
		DRCN (2016 ECCV)& $91.8$ & $73.7$ & $82.0$\\
		CoGAN (2016 NIPS)& $91.2$ & $89.1$ & $-$\\
		ADDA (2017 CVPR)& $89.4$ & $90.1$ & $76.0$\\
		UNIT (2017 NIPS)& $96.0$ & $93.6$ & $90.5$\\
		ATT (2017 ICML)& $-$ & $-$ & $86.2$\\
		GTA  (2019 CVPR)& $92.8$ & $90.8$ & $92.4$\\
		MSTN (2018 PMLR)& $92.9$ & $-$ & $91.7$\\
		DWT (2019 CVPR)& $99.1$ & $98.8$ & $97.8$\\
		DWT-MEC (2019 CVPR)& $99.0$ & ${\bf 99.2}$ & $97.8$\\
		ATM (2020 TPAMI)& $96.1$ & ${99.0}$ & $96.1$\\
		SAR (2020 IJCAI)& $94.1$ & ${98.0}$ & $-$\\
		\midrule
		Ours  & ${\bf 99.3}$ & ${\bf 99.2}$ & ${\bf 98.1}$ \\
		\bottomrule
	\end{tabular}
\vspace{-10pt}
\end{table}

\begin{table*}[ht!p]
	\centering
	    \setlength{\tabcolsep}{18pt}
	\caption{Classification accuracies (\%) on Office-31 dataset (ResNet50)}
	\label{tab:office-31}
	\footnotesize
	\vspace{-5pt}
	\begin{tabular}{lcccccc|cc}
		\toprule
		Method & A$\rightarrow$D & A$\rightarrow$W & D$\rightarrow$A & D$\rightarrow$W & W$\rightarrow$A & W$\rightarrow$D & Avg\\
		\midrule
		ResNet50& $68.9$ & $ 68.4$ & $62.5$ & $96.7$ & $60.7$ & $99.3$ & $76.1$\\ 
		GFK (2012 CVPR)& $74.5$ & $72.8$ & $63.4$ & $95.0$ & $61.0$ & $98.2$ & $77.5 $\\
		DAN (2015 ICML)& $78.6$ & $80.5$ & $63.6$ & $97.1$ & $62.8$ & $99.6$ & $81.4$ \\
		JAN (2017 ICML) & $84.7$ & $85.4$ & $68.6$ & $97.4$ & $70.0$ & $99.8$ & $84.3$ \\
		MADA (2018 AAAI)& $87.8$ & $90.0$ & $70.3$ & $97.4$ & $66.4$ & $99.6$ & $85.3$\\ 
		CAN (2018 CVPR)& $85.5$ & $81.5$& $65.9$ & ${ 98.2}$ & $68.0$ & ${\bf 100.0}$ & $86.5$\\ 
		CDAN (2018 NIPS)& ${ 89.8}$ & ${93.1}$& ${ 70.1}$ & $98.2$ & ${68.0}$ & ${\bf 100.0}$ & ${86.5}$ \\
		CDAN+E (2018 NIPS)& ${ 92.9}$ & ${\bf 94.1}$& ${ 71.0}$ & ${98.6}$ & ${ 69.3}$ & $98.0$ & ${87.3}$\\
		CRST (2019 ICCV)& ${ 88.7}$ & ${89.4}$& ${70.9}$ & ${98.9}$ & ${ 72.6}$ & ${\bf 100.0}$ & ${87.3}$\\
		BNM (2020 CVPR)& ${90.3}$ & ${91.5}$& ${ 70.9}$ & ${98.5}$ & ${71.6}$ & ${\bf 100.0}$ & ${87.1}$\\		  
		\midrule
		Ours  & ${\bf 95.5}$ & ${87.5}$ & ${\bf 74.7}$ & ${\bf 100.0 }$ & ${\bf 71.3}$ & ${\bf 100.0}$ & ${\bf 88.2}$ \\
		\bottomrule
	\end{tabular}
\end{table*}

\begin{table*}[ht!p]
	\centering
	\caption{Classification accuracies (\%) on Office-Home dataset (ResNet50)}
	\label{tab:office-home}
	\footnotesize
	\vspace{-5pt}
	\begin{tabular}{lcccccccccccc|cc}
		\toprule
		Method & P$\rightarrow$R & A$\rightarrow$P & C$\rightarrow$A & P$\rightarrow$C & C$\rightarrow$P & P$\rightarrow$A & A$\rightarrow$C  & P$\rightarrow$R & A$\rightarrow$R & R$\rightarrow$C & R$\rightarrow$A & C$\rightarrow$R &Avg\\
		\midrule
		ResNet50 & $59.9$ & $50$ & $37.4$ & $31.2$ & $41.9$ & $38.5$& $34.9$ & $31.2$  & $58.0$ & $41.2$ & $53.9$ & $46.2$ & $43.7$\\ 
		DAN (2015 ICML) & $74.3$ & $57.0$ & $45.8$ & $43.6$ & $56.5$ & $44.0$& $43.6$ & $67.7$  & $67.9$ & $51.5$ & $63.1$ & $60.4$ & $56.3$\\ 
		DANN (2016 JMLR)& $76.8$ & $59.3$ & $47.0$ & $43.7$ & $58.5$ & $46.1$& $45.6$ & $68.5$  & $70.1$ & $51.8$ & $63.2$ & $60.9$ & $57.6$\\ 
		
		JAN (2017 ICML)& $76.8$ & $61.2$ & $50.4$ & $43.4$ & $59.7$ & $45.8$& $45.9$ & $70.3$  & $68.9$ & $52.4$ & $63.9$ & $61.0$ & $58.3$\\ 
		
		CDAN-M (2018 NIPS)& $80.7$ & $65.9$ & $55.7$ & $49.1$ & $62.7$ & $51.8$& $50.6$ & $74.5$  & $73.4$ & $56.9$ & $68.2$ & $64.2$ & $62.8$\\ 
		
		CDAN (2018 NIPS)& $80.5$ & $69.3$ & $54.4$ & $48.3$ & $66.0$ & $55.6$& $49.0$ & $75.9$  & $74.5$ & $55.4$ & $68.4$ & $68.4$ & $63.8$\\ 
		
		CDAN+E (2018 NIPS)& ${\bf 81.6}$ & $70.6$ & $57.6$ & $50.9$ & $70.0$ & $57.4$& $50.7$ & $77.3$  & $76.0$ & $56.7$ & $70.9$ & $70.0$ & $65.8$\\ 
		DWT (2019 CVPR)& $78.2$ & $72.0$ & $58.9$ & $49.5$ & $65.6$ & $57.2$& $50.8$ & $78.3$  & $75.8$ & $55.3$ & $70.1$ & $60.2$ & $64.3$\\ 		
		ATM (2020 TPAMI)& $79.1$ & ${\bf 72.6}$ & $61.1$ & ${\bf 52.0}$ & ${\bf 72.0}$ & $59.5$& $52.4$ & $79.1$  & $78.0$ & ${\bf 58.9}$ & ${\bf 73.3}$ & ${\bf 72.6}$ & $67.9$\\ 			
		
		\midrule
		Ours & ${81.5}$ & ${\bf 72.6}$ & ${\bf 61.2}$ & ${51.3}$ & ${70.2}$ & ${\bf 60.0}$& ${\bf 56.2}$ & ${\bf 82.6}$  & ${\bf 79.2}$ & ${56.6}$ & ${\bf 73.3}$ & ${71.3}$ & ${\bf 68.0}$\\ 
		\bottomrule
	\end{tabular}
\end{table*}

\begin{table*}[ht!p]
	\centering
	    \setlength{\tabcolsep}{18pt}
	\caption{Classification accuracies (\%) on ImageCLEF-DA dataset (ResNet50)}
	\label{tab:imageclef-da}
	\footnotesize
	\vspace{-5pt}
	\begin{tabular}{lcccccc|cc}
		\toprule
		Method & I$\rightarrow$P & P$\rightarrow$I & I$\rightarrow$C & C$\rightarrow$I & C$\rightarrow$P & P$\rightarrow$C & Avg\\
		\midrule
		ResNet50 & $74.8$ & $ 83.9$ & $91.5$ & $78.0$ & $65.5$ & $91.2$ & $80.7$\\ 
		DAN (2015 ICML)  & $74.5$ & $82.2$ & $92.8$ & $86.3$ & $69.2$ & $89.8$ & $82.5 $\\
		DANN (2016 JMLR)& $75.0$ & $86.0$ & $96.2$ & $87.0$ & $74.3$ & $91.5$ & $85.0$ \\

		CAN (2018 CVPR)& ${78.2}$ & ${87.5}$& ${94.2}$ & $89.5$ & ${75.8}$ & $89.2$ & ${85.7}$\\  
		JAN (2017 ICML)& $76.8$ & $88.0$ & $94.7$ & $89.5$ & $74.2$ & $91.7$ & $85.8$ \\
		CDAN+E (2018 NIPS)& $77.7$ & $90.7$& $97.7$ & ${91.3}$ & $74.2$ & ${94.3}$ & $87.7$\\ 
		CDAN-M (2018 NIPS)& ${78.3}$ & ${91.2}$& ${96.7}$ & $91.2$ & ${77.2}$ & $93.7$ & ${88.1}$ \\

		iCAN (2018 CVPR) & ${79.5}$ & ${89.7}$& ${94.7}$ & $89.9$ & ${78.5}$ & $92.0$ & ${87.4}$\\  
		MEDA (2018 ACM MM)& ${80.2}$ & ${91.5}$& ${96.2}$ & $92.7$ & ${79.1}$ & $95.8$ & ${89.3}$\\
		ATM (2020 TPAMI)& ${80.3}$ & ${92.9}$& ${98.6}$ & $93.5$ & ${77.8}$ & $96.7$ & ${90.0}$\\
		DSAN (2020 TNNLS)& ${80.2}$ & ${93.3}$& ${97.2}$ & $93.8$ & ${80.8}$ & $95.9$ & ${90.2}$\\
		SAR (2020 IJCAI)& ${78.3}$ & ${91.3}$& ${96.7}$ & ${90.5}$ & ${78.1}$ & ${96.2}$ & ${88.5}$\\		  
		\midrule
		Ours  & ${\bf 84.7}$ & ${\bf 95.1}$ & ${\bf 99.4}$ & ${\bf 96.2 }$ & ${\bf 87.3}$ & ${\bf 98.0}$ & ${\bf 93.5}$ \\
		\bottomrule
	\end{tabular}
\end{table*}

\begin{table*}[ht!p]
	\centering

	\caption{Classification accuracies (\%) on VisDA-2017 dataset (ResNet101)}
	\label{tab:VisDA-C}
	\footnotesize
	\vspace{-5pt}
	\begin{tabular}{lcccccccccccc|cc}
		\toprule
		Method & aero & truck & train & skate & person & plant& motor & knife  & horse & car & bus & bicycle & Avg\\
		\midrule
		ResNet101 & $55.1$ & $8.5$ & $73.5$ & $26.5$ & $31.2$ & $81.0$& $79.7$ & $17.9$  & $80.6$ & $59.1$ & $61.9$ & $53.3$ & $52.4$\\ 
		DAN (2015 ICML) & $87.1$ & $20.7$ & $85.8$ & $36.3$ & $53.1$ & $49.7$& $63.0$ & $42.9$  & $90.3$ & $42.0$ & $76.5$ & $63.0$ & $59.2$\\ 
		DANN (2016 JMLR)& $81.9$ & $7.8$ & $82.8$ & $54.6$ & $65.1$ & $51.9$& $65.1$ & $29.5$  & $81.2$ & $44.3$ & $82.8$ & $77.7$ & $60.4$\\ 
		
		MCD (2018 CVPR)& $87.0$ & $25.8$ & $83.0$ & $40.3$ & $76.9$ & $88.6$& $84.7$ & $79.6$  & $88.9$ & $64.0$ & $83.7$ & $60.9$ & $72.0$\\ 
		
		BSP+DANN (2019 ICML) & $92.2$ & $37.1$ & $84.5$ & $66.9$ & $72.4$ & $80.6$& $86.8$ & $54.0$  & $87.0$ & $47.5$ & $83.8$ & $72.5$ & $72.1$\\ 
		
		CDAN (2018 NIPS)& $85.2$ & $38.0$ & $81.9$ & $76.0$ & $74.5$ & $83.4$& $88.1$ & $74.9$  & $84.2$ & $50.8$ & $83.0$ & $66.9$ & $74.0$\\ 
		
		BSP+CDAN (2019 ICML)& $92.4$ & $38.4$ & $82.1$ & ${\bf 77.9}$ & $77.0$ & $84.2$& $90.1$ & $80.6$  & $89.0$ & $57.5$ & $81.0$ & $61.0$ & $75.9$\\ 
		DSBN+MSTN (2019 CVPR)& $94.7$ & ${\bf 45.5}$ & ${88.3}$ & ${68.9}$ & ${\bf 81.3}$ & $91.1$& $87.9$ & $75.1$  & ${\bf 95.2}$ & $72.0$ & $76.0$ & ${\bf 86.7}$ & $80.2$\\ 		
		DSAN (2020 TNNLS)& $90.9$ & $39.4$ & ${\bf 89.1}$ & $67.6$ & $75.1$ & ${\bf 92.8}$& ${\bf 93.7}$ & $77.0$  & $88.9$ & $62.4$ & $75.7$ & $66.9$ & $75.1$\\ 		
		
		\midrule
		Ours & ${\bf 94.9}$ & $41.3$ & $87.8$ & $73.9$ & $77.8$ & ${91.2}$& ${93.2}$ & ${\bf 83.8}$  & $92.8$ & ${\bf 88.7}$ & ${\bf 86.5}$ & $76.5$ & ${\bf 82.4}$\\ 
		\bottomrule
	\end{tabular}
\end{table*}

\begin{table*}[ht!p]
	\centering
	    \setlength{\tabcolsep}{16pt}
	\caption{{Ablation study with regard to transferability on the Office-31 dataset}}
	\label{tab:ablation}
	\footnotesize
	\vspace{-5pt}
	\begin{tabular}{lcccccc|c}
		\toprule
		Methods& A$\rightarrow$D & A$\rightarrow$W & D$\rightarrow$A  & D$\rightarrow$W &W$\rightarrow$A&W$\rightarrow$D& Avg \\
		\midrule
		ResNet50& 68.9 & 68.4 & 62.5 & 96.7 &60.7&99.3& 66.1 \\
		+TDBN& 85.7 & 85.2 & 71.2 & 99.1 &68.9&{\bf 100.0}& 85.0 \\
		+OL& 88.8 & 83.2 & 72.3 & 99.0 &70.9&{\bf 100.0}& 85.7 \\
		\midrule
		Ours (TDBN+OL)  & {\bf 95.5}& {\bf 87.5}& {\bf 74.7}&{\bf 100.0}&{\bf 71.3}&{\bf 100.0}& {\bf 88.2}         \\
		\bottomrule
	\end{tabular}
\end{table*}

\subsection{Results}
\textbf{Office-31.}
Table~\ref{tab:office-31} quantifies the performance of ours employing TDBN and orthogonality weights and compares it with state-of-the-art records on Office-31 datasets. TDBN means that we replace the batch normalization layers with TDBN layers, and then we use the orthogonal weights in the fully-connected layers. In standard domain adaptation, our model outperforms such compared methods. Of them, most consider feature alignment without inter-domain relationships. From Table~\ref{tab:office-31}, our model behaves better than several compared methods, which account for the relationships within the features.

\textbf{Office-Home.} Table~\ref{tab:office-home} shows that ours exceeds the average accuracy more than 3.6\%. Also, considering Office-Home has 12 evaluations, from Table~\ref{tab:office-home}, we can easily find that our method achieves the best on all of them. Owing to Office-Home is a large-scale dataset, it can be concluded that our proposed method not only works well on standard benchmarks but also generalizes well on large-scale datasets.

{\bf ImageCLEF-DA.} The classification results of ImageCLEF-DA are shown in Table~\ref{tab:imageclef-da}. Our model outperforms the other compared methods on most transfer tasks. In particular, our overpasses DSAN by 3.3\%, which implies that redundancy reduction is beneficial for transferability.

{\bf VisDA-C.} Table~\ref{tab:VisDA-C} reports classification results of all the methods on the VisDA-C dataset. As Table 2 shows, the proposed UDA consistently achieves performance gains on all the adaptation tasks.

{\bf MNIST-USPS-SVHN.} We also consider the digit classification tasks on from MNIST to USPS and to SVHN. Experimental results are recorded in Table~\ref{tab:digit}. The proposed UDA model is superior to the other counterparts overally.

{\bf Overall Analysis.} The above experimental results reveal some insightful observations as follows: 1) In UDA, DWT \cite{roy2019unsupervised} directly utilizes a decorrelated batch normalization method and outperforms the previous methods, but it only considers to align global distribution. Differently, our method considers inter-domain and intra-domain features correlations as well as make feature standardization, which can capture more information on each category. 2) From object recognition, we can find ours achieves a great performance, which verifies the effectiveness of our model. 3) By comparing our method with several advanced GAN-based UDA methods, our method without extra hassle achieves the competitive performance with them, which shows its efficacy.

{\bf Ablation Study.} In this subsection, we will provide an ablation study analysis for our method. We perform ablative experiments on the Office-31 dataset to analyze the functionality of each component in our UDA model. Table~\ref{tab:ablation} summarizes the ablation results on Office-31 dataset using ResNet50 as the baseline, where the last column in the table reports the average recognition accuracy of ours as compared to two other configurations, i.e., TDBN without orthogonal loss, and orthogonal loss (OL) without TDBN. The comparative results directly show that both TDBN and OL greatly improve the performance of ResNet at a competitive level. Surprisingly, both cooperation in our model makes performance gains great. This implies that they are mutually complementary to each other. This is line with the previous claim that both schemes are conducted in a bi-level way: TDBN reduces the domain-specific feature redundancy, while OL does domain-invariant feature redundancy.

\begin{figure}[t]
    \vspace{0.1in}
    \centerline{\includegraphics[width=13cm]{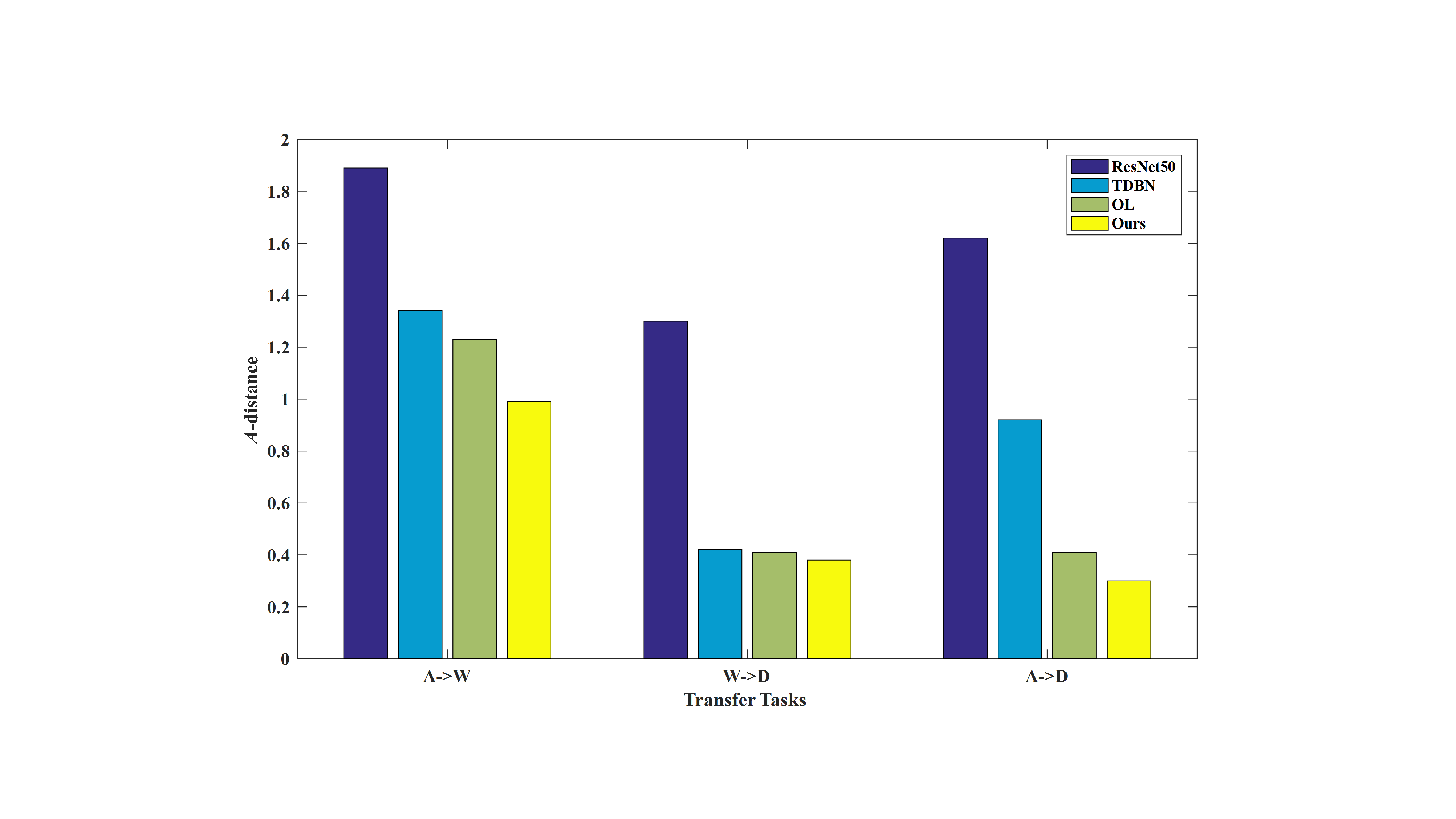}}
    \vspace{0.1in}
    \caption{The $\mathcal{A}$-distance across domain distributions on the Office-31 dataset}
    \label{distance}
\end{figure}

To further observe their ability to narrow the distribution gap across domains, the $\mathcal{A}$-distance \cite{ben2010theory} as the distribution discrepancy between source and target domain is used here to evaluate the functionality of each component of our model. Following \cite{ben2010theory}, the $\mathcal{A}$-distance is defined as $d_{\mathcal{A}}=2(1-2\epsilon)$, where $\epsilon$ is the classification error of a binary domain classifier for discriminating the source and target domains. We select all the transfer tasks on Office-31 to show the $\mathcal{A}$-distance bridged by ResNet-50, TDBN, OL, and our model, respectively, as in Fig.~\ref{distance}. From this figure, we can observe that simultaneous reducing inter-domain and intra-domain feature redundancy can greatly reduce the $\mathcal{A}$-distance across domains. 

\section{Conclusion}
In this paper, we propose a simple UDA approach from the perspective of feature redundancy minimization. Compared to existing arts that focus on feature discrimination, the proposed model seeks to reduce feature redundancy, thereby enhancing generalization the ability of the deep UDA model. Our model performs the reduction of \emph{bi-level} feature redundancy: the first level decorrelates domain-specific features, as well as matches feature distribution; the second level can stabilize the distribution of network activations, regularize deep neural networks and learn compact diverse features. Experiments on several large-scale publicly available image classification datasets show the superiority of our model against recent state-of-the-art traditional and deep counterparts.

\bibliographystyle{IEEEtran}
\bibliography{gUDA}

\appendix
\section{Basic Inequality Theorem}
{\bf Basic Inequality Theorem.} Any $n$ positive real number $x_{1}, x_{2}, \ldots, x_{n}$, there holds:
\begin{align} \label{eq:basic ineqality theorem}
{\frac{x_{1}+x_{2}+\ldots+x_{n}}{n} \geq \sqrt[n]{x_{1} \cdot x_{2} \cdots x_{n}}},
\end{align}
where the equality holds if and only if $x_{1}=x_{2}=\cdots=x_{n}$.

{\bf Proof} If $x_{1}, x_{2}, \ldots, x_{n}$ are positive real numbers, the mathematic average is 
$M=\frac{\sum_{1}^{n} x_{1}}{n}$ and the geometric average is $G=\left(\prod_{1}^{n} x_{i}\right)^{1 / n}$, then we can draw a basic conclusion that $M \geq G$.

Suppose $x_{1}, x_{2}, \ldots, x_{n}$ are arranged in an ascending order, when $n=1$, the inequality obviously holds. Suppose when $n=k$, then inequality holds. Then, when $n=k+1$, 
\begin{align} \label{eq:a.1}
{M_{k+1}=\frac{\sum_{1}^{k+1} x_{i}}{k+1}}.
\end{align}

Owing to the supposed sequence being ascending, we can get:
\begin{equation}
\label{eq:a.2}
\begin{array}{c}
\left(M_{k+1}-x_{1}\right)\left(x_{k+1}-M_{k+1}\right) \geq 0, \\
\left(x_{1}+x_{k+1}-M_{k+1}\right) M_{k+1} \geq x_{1} x_{k+1}.
\end{array} 
\end{equation}

According to Eq.~\eqref{eq:a.2}, we replace $x_{k+1}$ with $M_{k+1}$ and $x_{1}$ with $x_{1}+x_{k+1}-M_{k+1}$, respectively, to keep the arithmetic mean unchanged, then the new sequence is $\left(x_{1}+x_{k+1}-M_{k+1}\right), x_{2}, \ldots, x_{k}, M_{k+1}$, and its mean is still $M_{k+1}$. So the sub-sequence with the first $k$ element is:
\begin{equation}
\label{eq:a.3}
\begin{array}{c}
\left(x_{1}+x_{k+1}-M_{k+1}\right), x_{2}, \ldots, x_{k}. 
\end{array} 
\end{equation}

In terms of our hypothesis, the mean of \eqref{eq:a.3} is not larger than $M_{k+1}$, and then we can yield: 
\begin{equation}
\label{eq:a.4}
\begin{array}{c}
\left(x_{1}+x_{k+1}-M_{k+1}\right) * x_{2} * \ldots * x_{k} \leq\left(M_{k+1}\right)^{k}.
\end{array} 
\end{equation}

Let us multiply both sides by $M_{k+1}$. The Eq.~\eqref{eq:a.4} will become:
\begin{equation}
\label{eq:a.5}
\begin{array}{c}
\left(x_{1}+x_{k+1}-M_{k+1}\right) * x_{2} * \ldots * M_{k+1} \leq\left(M_{k+1}\right)^{k+1}.
\end{array} 
\end{equation}

According to Eq.~\eqref{eq:a.5},
\begin{equation}
\label{eq:a.5}
\begin{array}{c}
\left(M_{k+1}\right)^{k+1} \geq\left(x_{1}+x_{k+1}-M_{k+1}\right)* x_{2} * \ldots * x_{k} * M_{k+1} \\ ~~~~~~~
\geq x_{1} * x_{2} * \ldots * x_{k} * x_{k+1}=\left(G_{k+1}\right)^{k+1}.
\end{array} 
\end{equation}

Then, $M_{k+1} \geq G_{k+1}$. Obviously, the equality holds only when $x_{1}=x_{2}=\ldots=x_{n}$.
\begin{IEEEbiography}[{\includegraphics[width=1.2in,height=1.25in,clip,keepaspectratio]{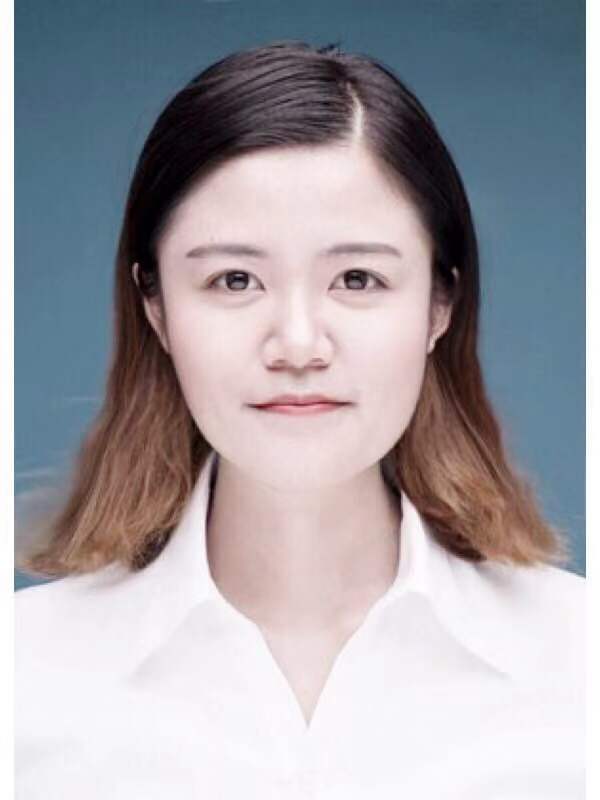}}]{Mengzhu Wang}
received a Bachelor's degree in Information and Computing Science from Tianjin University of Commerce,Tianjin, China, in 2016. She received the Master's degree from Chongqing University(CQU), China, in 2018. From 2019 to now, she is pursuing the Ph.D. degree with the School of Computer Science, the National University of Defense Technology, Changsha, China. Her current research interest include transfer learning, image segmentation and computer vision.
\end{IEEEbiography}

\begin{IEEEbiography}[{\includegraphics[width=1.2in,height=1.25in,clip,keepaspectratio]{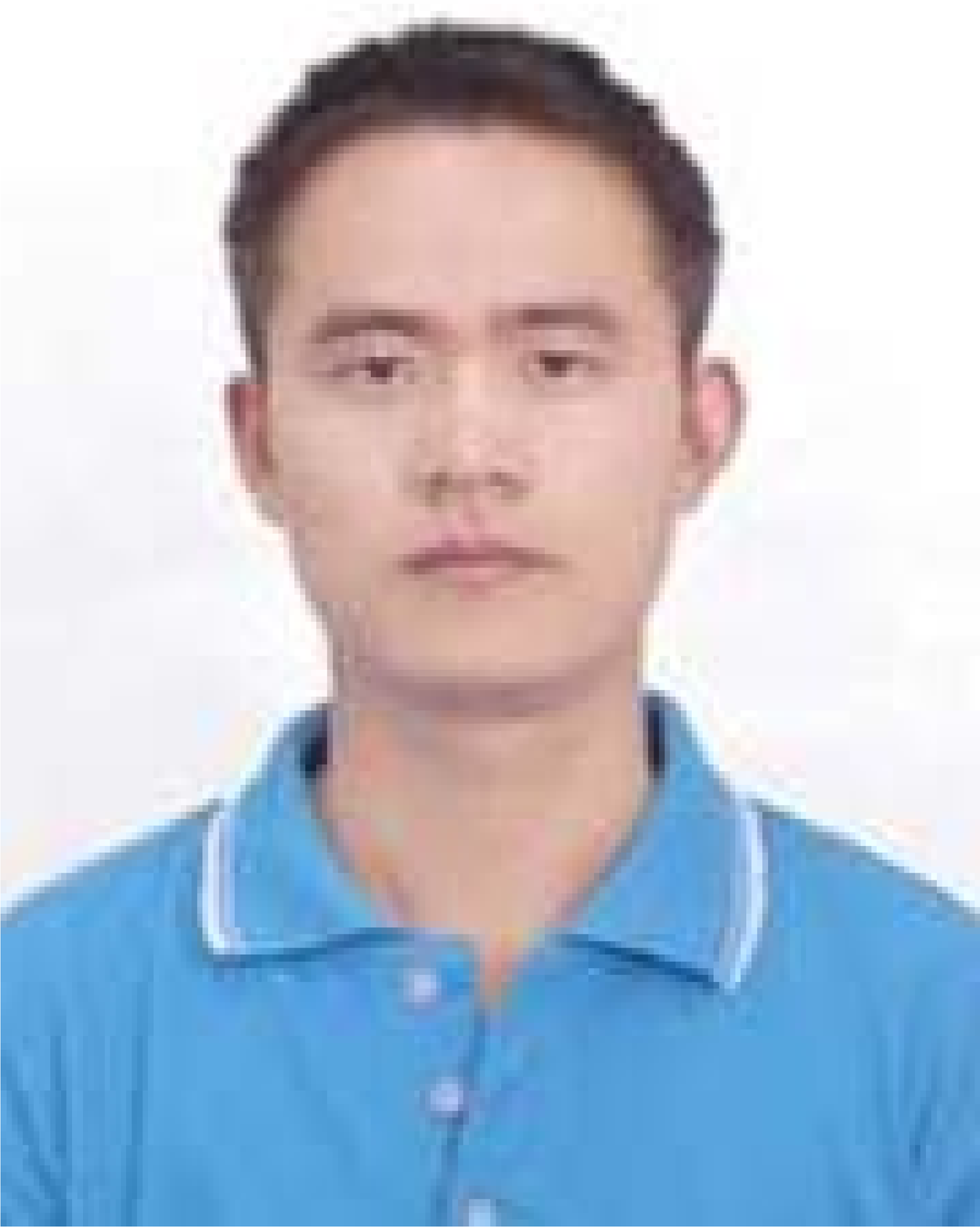}}]{Xiang Zhang} received the M.S., and Ph.D. degrees from the National University of Defense Technology, Changsha, China, in 2010 and 2015, respectively. He is currently a research assistant with the Institute for Quantum Information $\&$ State Key Laboratory of High Performance Computing, College of Computer, National University of Defense Technology. His current research interests include computer vision and machine learning.
\end{IEEEbiography}

\begin{IEEEbiography}[{\includegraphics[width=1.2in,height=1.25in,clip,keepaspectratio]{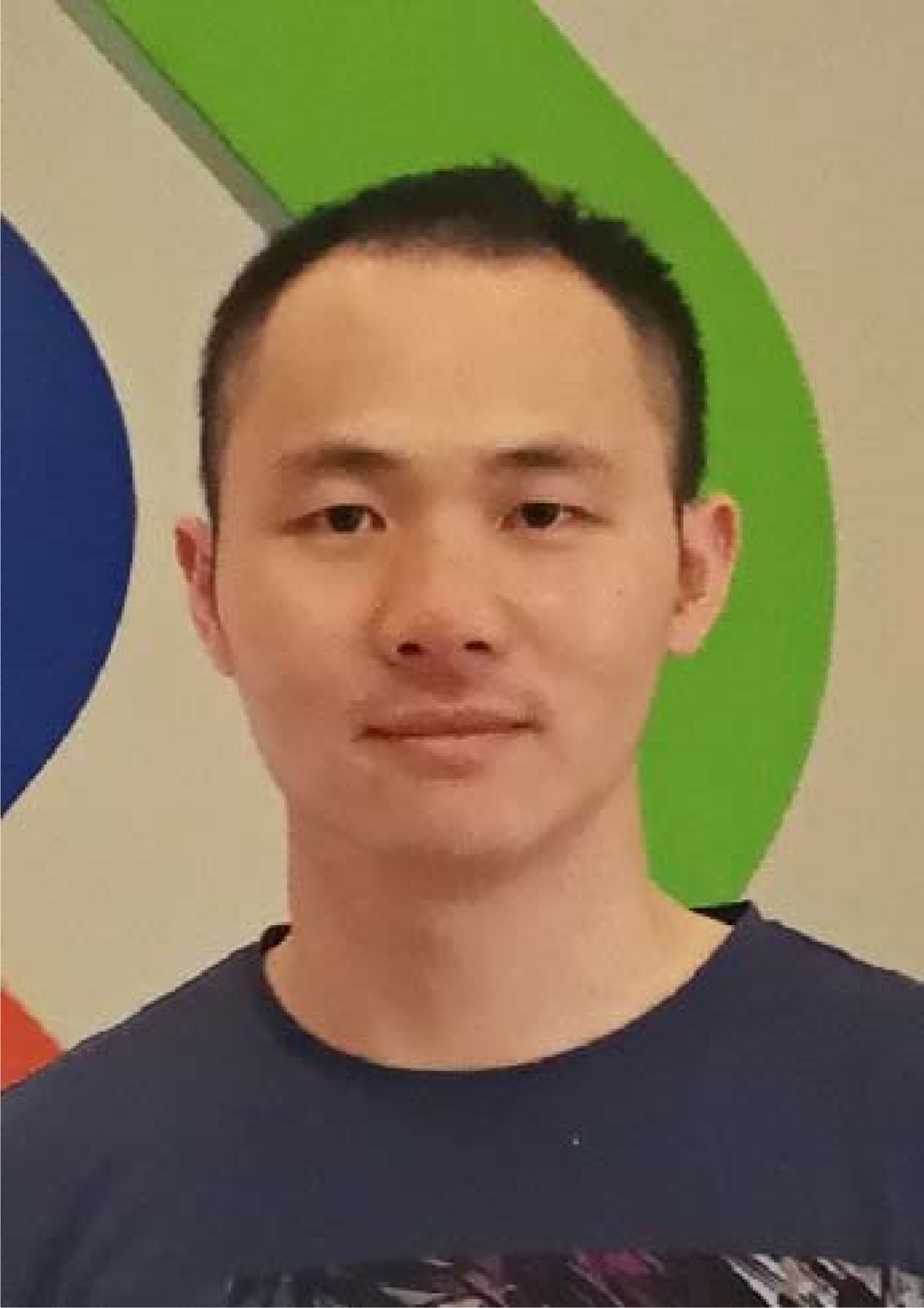}}]{Long Lan}
is currently a lecturer with College of Computer, National University of Defense Technology. He received the Ph.D. degree in computer science from National University of Defense Technology 2017. He was a visiting Ph.D. student in University of Technology, Sydney from 2015 to 2017. His research interests include multi-object tracking, computer vision and discrete optimization.
\end{IEEEbiography}

\begin{IEEEbiography}[{\includegraphics[width=1.2in,height=1.25in,clip,keepaspectratio]{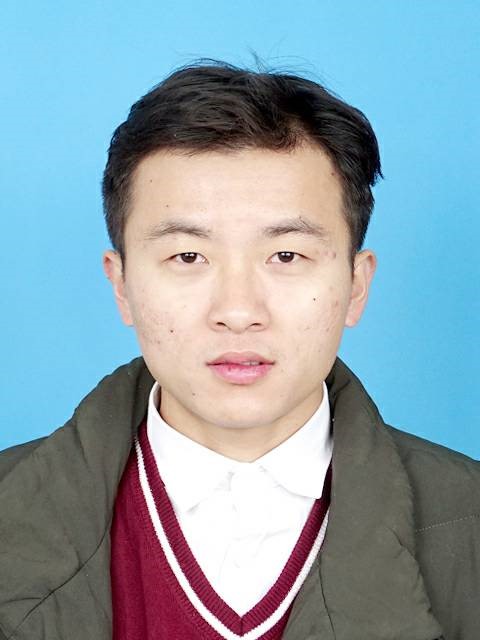}}]{Wei Wang}
	is currently a Ph.D. candidate at the School of Software Technology, Dalian University of Technology, Dalian, China. He received the B.S. degree at the school of science from the Anhui Agricultural University, Hefei, China, in 2015. He received the M.S. degree at the school of computer science and technology from the Anhui University, Hefei, China, in 2018. His major research interests include  transfer learning, zero-shot learning, deep learning, etc.
\end{IEEEbiography}

\begin{IEEEbiography}[{\includegraphics[width=1.2in,height=1.25in,clip,keepaspectratio]{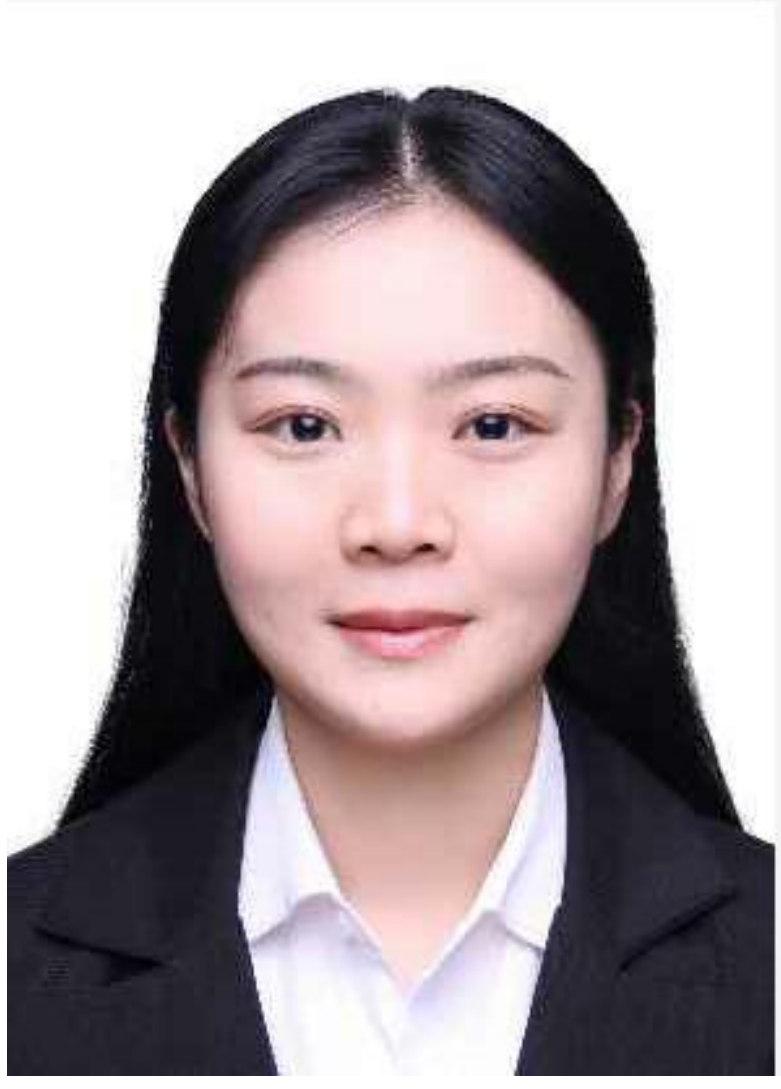}}]{Huibin Tan}
received a Bachelor's degree in Computer Science and Technology from Northeastern University of China in 2014. In 2014-2015, she studied postgraduate at the School of Computer Science, National University of Defense Technology, and later became a PhD student through a direct Ph.D application. From 2016 to now, she is pursuing the Ph.D. degree with the School of Computer Science, the National University of Defense Technology, Changsha, China. Her current research interest include face recognition, visual tracking and representation learning.
\end{IEEEbiography}

\begin{IEEEbiography}[{\includegraphics[width=1.2in,height=1.25in,clip,keepaspectratio]{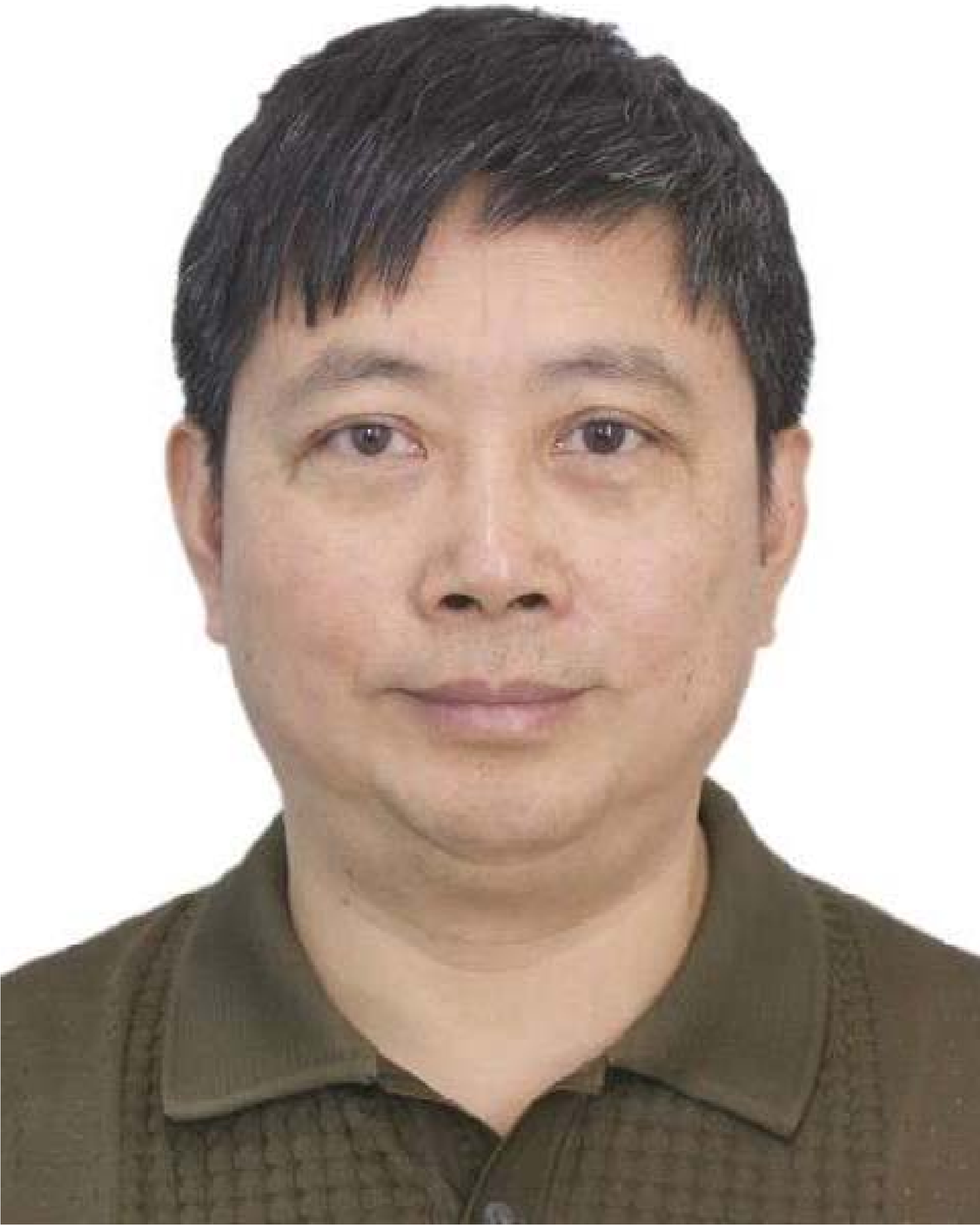}}]{Zhigang Luo}
received the B.S., M.S., and Ph.D. degrees from the National University of Defense Technology, Changsha, China, in 1981, 1993, and 2000, respectively. He is currently a Professor with the College of Computer, National University of Defense Technology. His current research interests include machine learning, computer vision and bioinformatics.
\end{IEEEbiography}  

\end{document}